\documentclass[runningheads]{llncs}
\usepackage{graphicx}
\usepackage{comment}
\usepackage{amsmath,amssymb} 
\usepackage{color}

\usepackage[ruled,vlined]{algorithm2e}
\usepackage{algorithmic}

\usepackage{multirow}
\usepackage{booktabs}
\usepackage[table,xcdraw]{xcolor}
\usepackage{ulem}
\usepackage{caption}
\usepackage{subcaption}
\usepackage{varwidth}
\usepackage{wrapfig}

\usepackage{pifont}
\newcommand{\cmark}{\ding{51}}%
\newcommand{\xmark}{\ding{55}}%

\newcommand{\etal}{{\it et al.}}
\newcommand{\eg}{{\it e.g.,}}
\newcommand{\ie}{{\it i.e.,}}
\newcommand{\wrt}{{\it w.r.t.}}

\newcommand{\figref}[1]{Figure~\ref{#1}}
\newcommand{\tabref}[1]{Table~\ref{#1}}
\newcommand{\algoref}[1]{Algorithm~\ref{#1}}

\newcommand\blfootnote[1]{%
  \begingroup
  \renewcommand\thefootnote{}\footnote{#1}%
  \addtocounter{footnote}{-1}%
  \endgroup
}

\begin{document}
\pagestyle{headings}
\mainmatter
\def\ECCVSubNumber{5930}  

\title{Fast Adaptation to Super-Resolution Networks via Meta-Learning} 

\titlerunning{Fast Adaptation to Super-Resolution Networks via Meta-Learning}
%

\author{Seobin Park*\inst{1} \and
Jinsu Yoo*\inst{1} \and
Donghyeon Cho\inst{2} \and
Jiwon Kim\inst{3} \\ \and
Tae Hyun Kim**\inst{1}}
\authorrunning{S. Park et al.}
%
\institute{Hanyang University, Seoul, South Korea, \email{\{seobinpark,jinsuyoo,taehyunkim\}@hanyang.ac.kr} \and
Chungnam National University, Daejeon, South Korea, \email{cdh12242@gmail.com} \and
SK T-Brain, Seoul, South Korea, \email{jk@sktbrain.com}}
\maketitle

\begin{abstract}

Conventional supervised super-resolution (SR) approaches are trained with massive external SR datasets but fail to exploit desirable properties of the given test image. On the other hand, self-supervised SR approaches utilize the internal information within a test image but suffer from computational complexity in run-time. In this work, we observe the opportunity for further improvement of the performance of single-image super-resolution (SISR) without changing the architecture of conventional SR networks by practically exploiting additional information given from the input image. In the training stage, we train the network via meta-learning; thus, the network can quickly adapt to any input image at test time. Then, in the test stage, parameters of this meta-learned network are rapidly fine-tuned with only a few iterations by only using the given low-resolution image. The adaptation at the test time takes full advantage of patch-recurrence property observed in natural images. Our method effectively handles unknown SR kernels and can be applied to any existing model. We demonstrate that the proposed model-agnostic approach consistently improves the performance of conventional SR networks on various benchmark SR datasets.
\keywords{Deep learning, Meta-learning, Single-image super-resolution, Patch recurrence}
\end{abstract}

\blfootnote{ * Equal contribution. ** Corresponding author.  }

\section{Introduction}

Super-resolution (SR) aims to increase the image size by recovering high-frequency details from a given low-resolution (LR) input image, and SR becomes a key feature in electrical goods, such as smartphone and TV; it has become popular as high-resolution (HR) screens are commonly available in our daily lives. The most basic methods utilize interpolation techniques (\eg nearest and bicubic resizing) to fill in the missing pixels. These methods are efficient but produce blurry results. Moreover, dedicated hardware-equipped devices, such as jittering~\cite{jitter} and focal stack~\cite{focal_stack} cameras, allow the use of multiple images to solve the LR image problem. However, these specialized devices incur additional costs and cannot be used to restore images captured with conventional cameras in the past. To mitigate these problems, numerous single-image super-resolution (SISR) algorithms that restore high-quality images by using only a single LR image as input have been studied; in particular, optimization-based~\cite{glasner,selfex} and learning-based~\cite{srcnn,vdsr,enet,idn,zssr,drrn,memnet} methods have been investigated intensively.


Since Dong~\etal~\cite{srcnn} demonstrated that a three-layered convolutional neural network could outperform the traditional optimization-based methods by a large margin, researchers have proposed numerous deep-learning-based methods for SISR. These methods aim to increase the performance of peak signal-to-noise ratio (PSNR) and structural similarity (SSIM) by allowing deeper networks to maximize the power of deep learning with large training datasets. 
In recent years, however, PSNR values have reached a certain limit, and more studies using perception metric~\cite{johnson,srgan} have been introduced to focus on creating visually pleasing and human-friendly images.

Most of the current deep-supervised-learning approaches do not explicitly adapt their models during test time. Instead, fixed network parameters are used for all test images regardless of what we can learn more from the new test image. To fully utilize the additional information available from the given input test image (LR), we propose to extend this single fixed model approach by combining it with a dynamic parameter adaptation scheme. We find that the adaptive network results in better performance, especially for unseen type of images. In particular, we can utilize patch-recurrence property if available in the input image, which can be described as self-supervised learning. The notion of exploiting patch-recurrence has been introduced in prior works \cite{glasner,zontak2011internal}. Recently, Shocher \etal~\cite{zssr} proposed a zero-shot SR (ZSSR) method employing deep learning; this study is the most related work to our proposed method. ZSSR trains a relatively small convolutional neural network at test time from scratch, with training samples extracted only from the given input image itself. Therefore, ZSSR can naturally exploit the internal information of the input image. However, ZSSR has some limitations: (1) requirement of considerable inference time due to slow self-training step; (2) failure to take full advantage of using pre-trained networks learned by large amounts of external dataset;  

\begin{table}[t]
\centering
\begin{tabular}{@{}lccc@{}}
\toprule
& \multicolumn{1}{l}{~~External dataset?~~} & \multicolumn{1}{l}{~~Internal dataset?} & \multicolumn{1}{l}{~~Fast at run-time?~~} \\ \midrule
Self-supervision~\cite{zssr}    & \xmark & \cmark & \xmark             \\
Supervision~\cite{srcnn,enet,idn}  & \cmark  & \xmark  & \cmark   \\
\textbf{MLSR(ours)} & \cmark & \cmark & \cmark \\ \bottomrule
\end{tabular}
\caption{Conventional supervised SR methods are trained with external SR datasets and run fast. Whereas, self-supervised SR methods typically exploit information using the given test image at run-time, which is time-consuming and impractical. Meta-learning for SR (MLSR) can efficiently utilize both external and internal information and take advantages of each approach. 
}
\label{table_intro}
\end{table}

Meta-learning can be a breakthrough for the above-mentioned problem. Meta-learning, \ie learning to learn, is gaining popularity in recent deep-learning studies~\cite{learningtolearn,prototypical}. Meta-learning aims to learn quickly and efficiently from a small set of data available at test time. Several methods, such as recurrent architecture-based~\cite{ntm}, and gradient-based methods~\cite{mlgradient,fomaml}, have been proposed. In particular, model agnostic meta-learning (MAML)~\cite{maml} is an example of a gradient-based method. We experimentally find that training a network with MAML results in the best initialization of the network parameter to perform well when fine-tuning with a small number of given input data.
Consequently, as shown in~\tabref{table_intro}, the proposed method can efficiently utilize both external and internal information and take advantages of each approach.

To this end, we introduce a method employing the meta-learning (fast adaptation) algorithm to solve the SISR problem. Using a large number of degraded images generated with various SR kernels, our SR networks are trained not only to generalize over large external data but also to adapt fast to any input image with real SR kernels.

Our contributions can be summarized as follows:
\begin{itemize}

\item To our knowledge, fully exploiting supervision signals available from both external and internal data with an effective meta-learning method is successful for the first time. 
\item Most state-of-the-art SR networks can be improved with our meta-learning scheme without changing the predefined network architectures.
\item Our method achieves the state-of-the-art performance over benchmark datasets.

\end{itemize}

\section{Related Works}
In this section, we review the most relevant SISR works. Also, methods for handling unknown SR kernel (\ie blind SR) are briefly described.

An early example-based SISR method~\cite{freeman2002example} learned the complicated relationships between LR and HR patches by learning how to use the external dataset. A locally linear embedding-based SISR method was introduced by Chang~\etal~\cite{lle}. Yang~\etal~\cite{sparse_code} proposed a sparse coding-based algorithm assuming that a pair of LR and HR patches shares the same sparse coefficients with each distinct dictionary. Also, learning methodologies like random forest~\cite{random_forest1,random_forest2,random_forest3}, hierarchical decision tree~\cite{tree}, and deep learning ~\cite{srcnn,vdsr,enet,srgan,idn} have been proposed to boost the performance of SISR.

%
The self-similarity-based methods assume that a natural image contains repetitive patterns and structures within and across different image scales. Glasner~\etal~\cite{glasner} proposed a unified framework that incorporates self-similarity-based approaches by exploiting patch-recurrence within and across different scales of a given LR input image. Huang~\etal~\cite{selfex} handled transformed patches by estimating the transformations between the corresponding LR\textendash HR patch pairs. Dong~\etal~\cite{nonlocal_sparse} proposed the non-local centralized sparse representation to exploit the non-local self-similarity of a given LR image. Huang~\etal~\cite{SRHRF+} combined the benefits from both external and internal databases for SISR by employing a hierarchical random forest.

Recently, a study dealing with the unknown SR kernel has begun to draw attention. Michaeli and Irani~\cite{blindsr} exploit the nature of recurrence of small patches to handle unknown SR kernel. Yuan~\etal~\cite{unsupervisedis} propose an unsupervised learning method with CycleGAN~\cite{zhu2017unpaired}, and Gu~\etal~\cite{blindsr2019} estimate unknown SR kernel iteratively and additionally add a spatial feature transform (SFT) layers into the SR network for handling multiple blur kernels. Based on a simple convolutional neural network, ZSSR~\cite{zssr} deals with SR kernels given at test time by exploiting information from an input image itself.

In this work, we focus on overcoming the limitations of these conventional SISR methods. We observe that many existing deep learning-based methods fail to fully utilize the information provided in a given input image. Although ZSSR~\cite{zssr} utilizes both the power of deep learning and information from the input image at test time, it does not use pre-trained networks with large external dataset. Therefore, we start from training network to utilize the external examples. Then, we fine-tune the network with the input image to utilize the information captured by internal patch-recurrence and cover unknown SR kernels (given at test time) for the input. To obtain a well-trained network that can quickly adapt to the input image by using patch-recurrence, we integrate a meta-learning technique inspired by MAML~\cite{maml} with conventional SISR networks without changing the network architectures. MAML aims to learn a meta-network that can swiftly adapt to new learning task and examples using a small number of iterations for fine-tuning. MAML is applicable in a variety of tasks, such as few-shot learning~\cite{maml_fewshot1,maml_fewshot2} and reinforcement learning~\cite{maml_rl}. We apply the MAML method to fine-tune the pre-trained SR parameters to input images quickly and efficiently. We experimentally verify that our approach can boost the performance of state-of-the-art SISR by a large margin.

\section{Meta-Learning for Super-Resolution}

In this section, we introduce a new neural approach that integrates recent meta-learning techniques to solve the SISR problem by exploiting additional information available in the input LR image. 

\subsection{Exploiting patch-recurrence for deep SR}
\label{sec_patch_recurrence}

According to~\cite{glasner,selfex,zssr}, small patches inside the natural image recur multiple times across the different scales of a given image. Therefore, unlike conventional learning-based methods that utilize large external datasets, we can find multiple HR patches corresponding to a given LR patch within a single-input image using the patch-recurrence. However, these previous methods have been developed separately and handle external and internal datasets differently. Thus, we develop a new method that facilitates SR networks by utilizing both (large) external and (small) internal datasets to further enhance the quality of the restored images.

\begin{figure}[t]
\centering
\includegraphics[width=\linewidth]{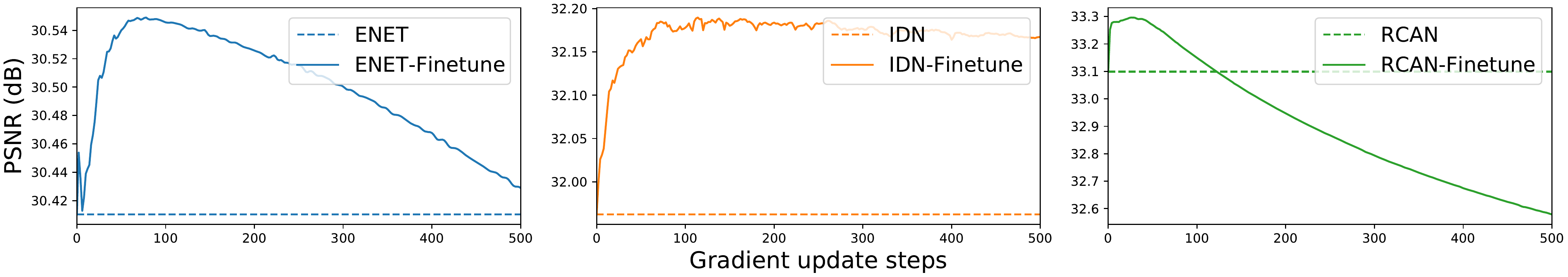}

\caption{Increasing PSNR values of ENET~\cite{enet}, IDN~\cite{idn}, and RCAN~\cite{rcan} with fine-tuning process during the testing phase.}
\label{fig_intro_finetune}

\end{figure}

First, we conduct a simple experiment to improve the performance of existing deep SR networks without changing their network architectures by using the patch-recurrence from a given LR test image. To achieve such goal, we re-train (fine-tune) the fully trained SR networks, such as ENET~\cite{enet}, IDN~\cite{idn}, and RCAN~\cite{rcan}, with a new training set consisting of LR test image and its down-scaled version ($\times0.5$). Note that, RCAN is currently state-of-the-art SR network. By updating the network parameters using the gradient descent, the PSNR values of SR networks increase. Also note that we can further increase PSNR values without using the ground truth HR image, because we utilize additional information obtained from the patch-recurrence of the new training set (\figref{fig_intro_finetune}). The PSNR values in~\figref{fig_intro_finetune} are obtained by calculating the average of the updated PSNR values on the Urban100 dataset~\cite{selfex}. The PSNR values tend to increase until 50 iterations, then decrease because the networks can be over-fitted with a small training set at test time. The performance of RCAN drops relatively quickly due to huge number of parameters used in the network.

This experiment demonstrates that there is still room to improve the performance of conventional SR networks while keeping their original network architectures, and patch-recurrence property is a key to boost the performance by adapting parameters of the fully pre-trained networks.

\subsection{Handling unknown SR kernel for deep SR}
\label{sec_unknown}

SR in unknown degradation settings (\ie unknown SR kernel) is more challenging than conventional SR problem in ideal setting using the bicubic interpolation. 
According to~\cite{blindsr}, the performance of the conventional SR networks trained with only bicubic kernel deteriorates significantly when it comes to the non-bicubic and real SR kernels in real scenario~\cite{zssr}. That is, generalization capability of the networks which can handle newly seen SR kernel during test phase is restricted in real situation. However, many conventional SR networks still assume ideal and fixed bicubic SR kernel, and thus cannot handle real non-bicubic SR kernels.

\begin{table}[t]
\centering
\begin{tabular}{lcccc}
\hline
Model & ~~Set5~~ & ~~Set14~~ & ~~BSD100~~ & ~~Urban100~~ \\ \hline
IDN~\cite{idn} & 28.24 & 26.29 & 26.28 & 23.49 \\
IDN-Finetune~~~ & \textbf{31.52} & \textbf{28.91} & \textbf{28.47} & \textbf{25.93} \\ \hline
\end{tabular}
\caption{Average PSNR for scale factor $\times$2 dealing with unknown SR kernel.}
\label{table_finetune_unknown}
\end{table}

In this section, we perform a simple experiment to see whether this problem can be also alleviated with patch-recurrence property.
We first degrade input LR image with a non-bicubic SR kernel to generate a new training set consisting of LR image and its down-sized image ($\times 0.5$), and then evaluate the performance of the original IDN and its fine-tuned version with the down-sized training set. Note that the original IDN is initially trained with only bicubic SR kernel on the DIV2K~\cite{DIV2K} dataset, and our IDN with fine-tuning (IDN-Finetune) is further optimized for 1000 iterations with the gradient update. 
In~\tabref{table_finetune_unknown}, we observe that IDN-Finetune can handle the images degraded by non-bicubic SR kernel much better on numerous benchmark datasets compared to the original IDN trained with bicubic SR kernel. 
Thus, we can see that the patch-recurrence property still holds and can be also used to improve SR performance by handling unknown SR kernels.

\subsection{Proposed Method}
\label{sec_proposed_method}

In previous sections, we have shown that the patch-recurrence property can be used not only to improve the performance of SR networks but also to deal with non-bicubic SR kernels. However, to update and adapt the pre-trained network parameters at test time to the specific input image, naive fine-tune-based update with stochastic gradient descent (SGD) requires large number of iterations and takes much time. To solve this problem, we integrate a meta-learning technique~\cite{maml} with the SR networks to facilitate use of the patch-recurrence and boost the speed of the adaptation procedure at test time.

First, we define each task to employ meta-learning as super-resolving a single specific LR image by utilizing internal information available within the given LR input image.
However, unlike conventional few-shot/k-shot problems which can be solved by meta-learning, our new SR task does not provide the ground-truth data (HR image) corresponding to the LR input image for adaptation at test time. Thus, it is difficult to directly apply the conventional meta-learning algorithms to our new learning task for SR.

Therefore, we develop a Meta-Learning for SR (MLSR) algorithm based on our observation that a pair of images composed of LR input and its down-scaled version (LR$\downarrow$) can be used as a new training sample for our new SR task due to the patch-recurrence property of the natural image, which learns to adapt the pre-trained SR networks to the given test image.
To be specific, we employ the recent model-agnostic meta-learning (MAML) approach. In particular,
MAML allows fast adaptation to a new task with only a small number of gradient updates~\cite{maml}, so we can boost the speed of our test-time learning task which originally requires large number of gradient update steps without meta-learning scheme (\ie \space naive fine-tune).

\begin{figure*}[t]
\centering
\includegraphics[width=1.0\linewidth]{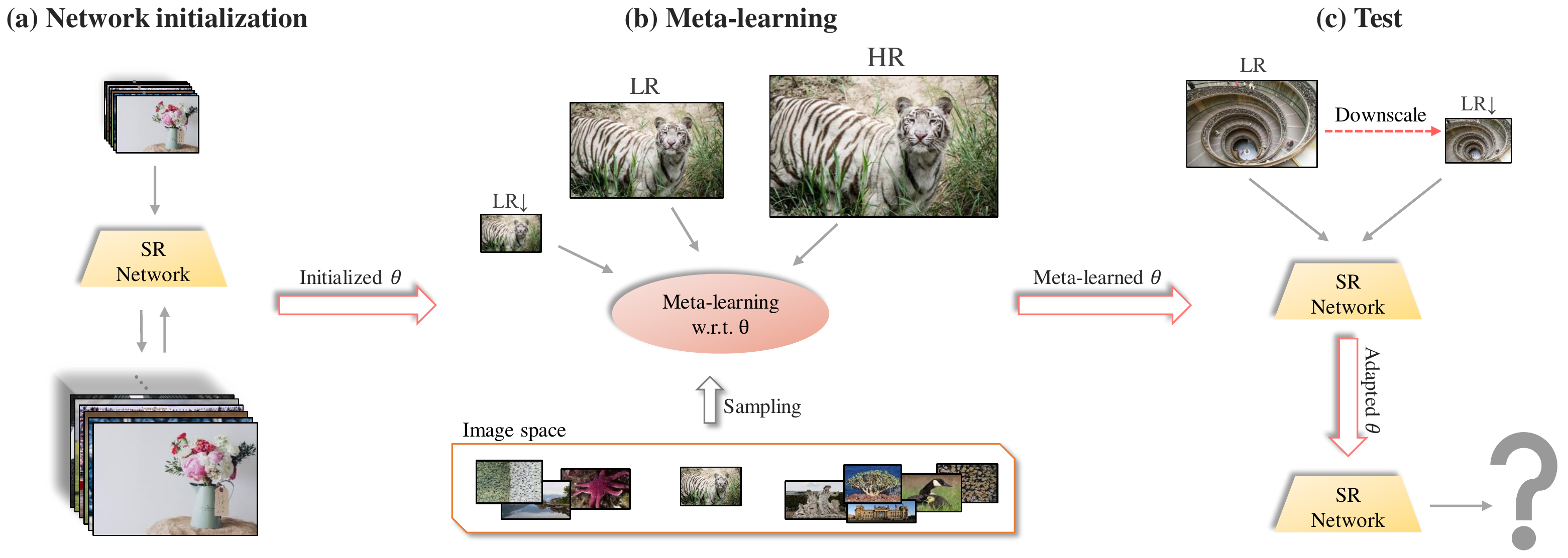}
\caption{Overall flow of the proposed method (MLSR). (a) Initialization stage of MLSR. Conventional SR network is trained with large external dataset. (b) Meta-learning stage of MLSR. The SR network is meta-trained to allow fast adaptation to any input image at test time. (c) Test stage of MLSR. Meta-learned parameters are rapidly tuned to the given LR image. }
\label{overall_flow}
\end{figure*}

In~\figref{overall_flow}, the overall flow of the proposed method is illustrated. 
First, we initialize the conventional SR networks with large external train datasets.
Next, we start meta-learning using MAML which optimizes the initialized SR parameters to enable quick adaptation to the new LR test image.
Finally, during the test phase, we adapt the meta-learned parameters with the given LR test image, and restore the HR image by using the adapted parameters.

\normalem
\begin{algorithm}[b]

 \algorithmicrequire \space $p(I)$: Distribution (\eg uniform) over images \\
 \algorithmicrequire \space $\alpha$, $\beta$: Hyper-parameters (step-size)\newline \\
 
 \nl Initialize $\theta$ \\
 \nl \While{not converged}{
 \nl Sample a batch of images $\{I_i\} \sim p(I)$ \\
 \nl Generate $\{HR_i\}$, $\{LR_i\}$, $\{LR_i\downarrow\}$ from $\{I_i\}$ \\

 \nl \ForEach{$i$}{
 \nl Evaluate $\nabla_\theta\mathcal{L}(f_\theta(LR_i\downarrow), LR_i)$ using $\mathcal{L}$ \\
 \nl Compute adapted parameters with SGD: $\theta_i \leftarrow \theta - \alpha\nabla_\theta\mathcal{L}(f_\theta(LR_i\downarrow), LR_i)$ \\
  }

 \nl Update $\theta \leftarrow \theta - \beta\nabla_\theta\sum_{i}\mathcal{L}(f_{\theta_i}(LR_i), HR_i)$
 }
 
 \caption{MLSR training algorithm}
 \label{algorithm_mlsr}

\end{algorithm}

\begin{algorithm}[t]
 \algorithmicrequire \space $I$: Given image \\
 \algorithmicrequire \space $\alpha$: Hyper-parameter (step-size) \\
 \algorithmicrequire \space $n$: Number of gradient updates \newline \\
 \nl Initialize $\theta$ with meta-trained parameter \\
 \nl Generate $LR$, $LR\downarrow$ from $I$ \\ 
  \nl $i \leftarrow 0$ \\
 \nl \While{i $<$ n}{
 \nl Compute adapted parameters with SGD: $\theta \leftarrow \theta - \alpha\nabla_\theta\mathcal{L}(f_\theta(LR\downarrow), LR)$ \\
 \nl $i \leftarrow i+1$
 }
 \nl Compute $f_{\theta}(LR)$

 \caption{MLSR inference algorithm}
 \label{algorithm_mlsr_test}
\end{algorithm}
\ULforem

Specifically, we formulate the proposed method more concretely.
Our SR model $f_{\theta}$ which is initialized with parameter $\theta$ renders an HR image from a given LR image by minimizing the loss $\mathcal{L}$, and it yields,
\begin{equation}
\mathcal{L}(f_\theta(LR), HR) = ||f_\theta(LR) - HR||_{2}^{2},
\label{loss}
\end{equation}
and our goal of meta-learning is to optimize the network parameter $\theta$ to be quickly adapted to $\theta_i$ at test time with the given input image $LR_i$ and its down-scaled image $LR_i\downarrow$.
Therefore, the adaptation formulation with gradient update is given as follows: 


\begin{equation}
\theta_i = \theta - \alpha\nabla_\theta\mathcal{L}(f_\theta(LR_i\downarrow), LR_i),
\label{eq_update_inner}
\end{equation}
where hyper-parameter $\alpha$ controls the learning rate of the inner update procedure.
Notably, to generate the down-scaled image $LR\downarrow$ we can use any SR kernel if available.
Then, we optimize the following objective function  \wrt \space $\theta$:
\begin{equation}
\operatorname*{argmin}_{\theta} \sum_{i}\mathcal{L}(f_{\theta_i}(LR_i), HR_i),
\label{eq_objective}
\end{equation}
and the optimization is preformed by the gradient update as: 
\begin{equation}
\theta \leftarrow \theta - \beta\nabla_\theta\sum_{i}\mathcal{L}(f_{\theta_i}(LR_i), HR_i).
\label{eq_outter}
\end{equation}

In general, we can use multiple iterations for the adaptation in \eqref{eq_update_inner},
but it increases computational cost in calculation of high-order derivatives in \eqref{eq_outter}.
To alleviate this problem, we can simply employ the first-order approximation methods~\cite{maml,fomaml},
which is known to give competitive results with lower computational cost. 
In our experiments, we use the first-order MAML introduced in~\cite{maml}. 



At test time, we first adapt the parameters ($\theta$) of the meta-learned SR network with the input LR image ($LR$) and its down-sized image ($LR\downarrow$), then restore the HR image using the adapted SR parameters as elaborated in ~\algoref{algorithm_mlsr_test}.

\section{Experimental Results}

In this section, we perform extensive experiments to demonstrate the superiority of the proposed method, and show quantitative and qualitative comparison results. Our source code is publicly available.\footnote[1]{\url{https://github.com/parkseobin/MLSR}}

\subsection{Implementation details}%

For our experiments, we first pre-train conventional SR networks (SRCNN~\cite{srcnn}, ENET~\cite{enet}, IDN~\cite{idn}, and RCAN~\cite{rcan}) with DIV2K~\cite{DIV2K} dataset. We use publicly available pre-trained parameters for IDN and RCAN (TensorFlow versions), and use our own parameters trained from scratch for SRCNN and ENET.
Next, we start meta-learning for these baseline SR networks in accordance with iterative steps in~\algoref{algorithm_mlsr}. For meta-learning, we still use DIV2K dataset, and use 5 inner gradient update steps in \eqref{eq_update_inner} (line 7 in \algoref{algorithm_mlsr}). We set $\alpha=10^{-5}$, $\beta=10^{-6}$, train patch size to 512$\times$512, and mini-batch size to 16.

\subsection{MLSR with fixed bicubic SR kernel}

\begingroup
\setlength{\tabcolsep}{4.5pt}
\renewcommand{\arraystretch}{1.2}
\begin{table}[t]
\begin{tabular}{lccccccc}
\hline
\multirow{2}{*}{Model} & \multirow{2}{*}{Iteration} & \multicolumn{2}{c}{DIV2K} & \multicolumn{2}{c}{BSD100} & \multicolumn{2}{c}{Urban100} \\ \cline{3-8} 
 &  & PSNR & SSIM & PSNR & SSIM & PSNR & SSIM \\ \hline
SRCNN~\cite{srcnn} & - & 34.11 & 0.9272 & 31.13 & 0.8852 & 29.39 & 0.8927 \\
\multirow{3}{*}{\textbf{+ MLSR (ours)}} & 5 & 34.14 & 0.9274 & 31.15 & 0.8855 & 29.42 & 0.8931 \\
 & 20 & 34.18 & 0.9276 & 31.19 & 0.8857 & 29.48 & 0.8936 \\
 & 100 & \textbf{34.23} & \textbf{0.9281} & \textbf{31.22} & \textbf{0.8860} & \textbf{29.54} & \textbf{0.8945} \\ \hline
ENET~\cite{enet} & - & 34.59 & 0.9329 & 31.64 & 0.8935 & 30.38 & 0.9097 \\
\multirow{3}{*}{\textbf{+ MLSR (ours)}} & 5 & 34.62 & 0.9331 & \textbf{31.69} & \textbf{0.8936} & 30.46 & 0.9105 \\
 & 20 & 34.64 & 0.9333 & \textbf{31.69} & \textbf{0.8936} & 30.49 & 0.9108 \\
 & 100 & \textbf{34.67} & \textbf{0.9335} & 31.67 & 0.8934 & \textbf{30.52} & \textbf{0.9112} \\ \hline
IDN~\cite{idn} & - & 35.24 & 0.9403 & 32.11 & 0.8994 & 31.95 & 0.9269 \\
\multirow{3}{*}{\textbf{+ MLSR (ours)}} & 5 & 35.36 & 0.9408 & \textbf{32.17} & \textbf{0.8996} & 32.06 & 0.9275 \\
 & 20 & 35.38 & 0.9409 & \textbf{32.17} & \textbf{0.8996} & 32.17 & 0.9280 \\
 & 100 & \textbf{35.40} & \textbf{0.9413} & 32.08 & 0.8988 & \textbf{32.23} & \textbf{0.9286} \\ \hline
RCAN~\cite{rcan} & - & 35.69 & 0.9451 & 32.38 & \textbf{0.9023} & 33.10 & 0.9369 \\
\multirow{3}{*}{\textbf{+ MLSR (ours)}} & 5 & 35.72 & 0.9454 & \textbf{32.39} & \textbf{0.9023} & 33.27 & 0.9373 \\
 & 20 & \textbf{35.75} & \textbf{0.9458} & 32.37 & 0.9022 & \textbf{33.32} & \textbf{0.9379} \\
 & 100 & 35.48 & 0.9444 & 32.04 & 0.8982 & 33.26 & 0.9373 \\ \hline
\end{tabular}
\caption{PSNR and SSIM results from different SR networks on different test dataset with scale $\times$2. Bicubic SR kernel is used. Baseline SR networks are SRCNN~\cite{srcnn}, ENET~\cite{enet}, IDN~\cite{idn}, and RCAN~\cite{rcan}, and \textbf{+ MLSR} indicates the meta-learned version of the baseline network.}
\label{table_mlsr_overall}
\end{table}
\endgroup

\begin{table}
\begingroup
\renewcommand{\arraystretch}{1.3}
\setlength{\tabcolsep}{.9ex}
\begin{varwidth}[c]{0.55\linewidth}
\centering
\begin{tabular}{llc}
\hline
Dataset & Model & Before/after adaptation \\ \hline
\multirow{2}{*}{DIV2K} & IDN~\cite{idn} & 32.19 / - \\
 & IDN-ML & 32.19 / \textbf{32.37} \\ \hline
\multirow{2}{*}{Urban100*} & IDN~\cite{idn} & 32.28 / - \\
 & IDN-ML & 32.13 / \textbf{32.43} \\ \hline
\end{tabular}
\caption{PSNR results of IDN~\cite{idn} and IDN-ML trained on different datasets. 40 images in Urban100 dataset are selected for the evaluation. * Another 50 images in the dataset are used for training.}
\label{table_urban}
\end{varwidth}
\endgroup
\hfill
\begin{minipage}[c]{.4\linewidth}
\includegraphics[width=\linewidth]{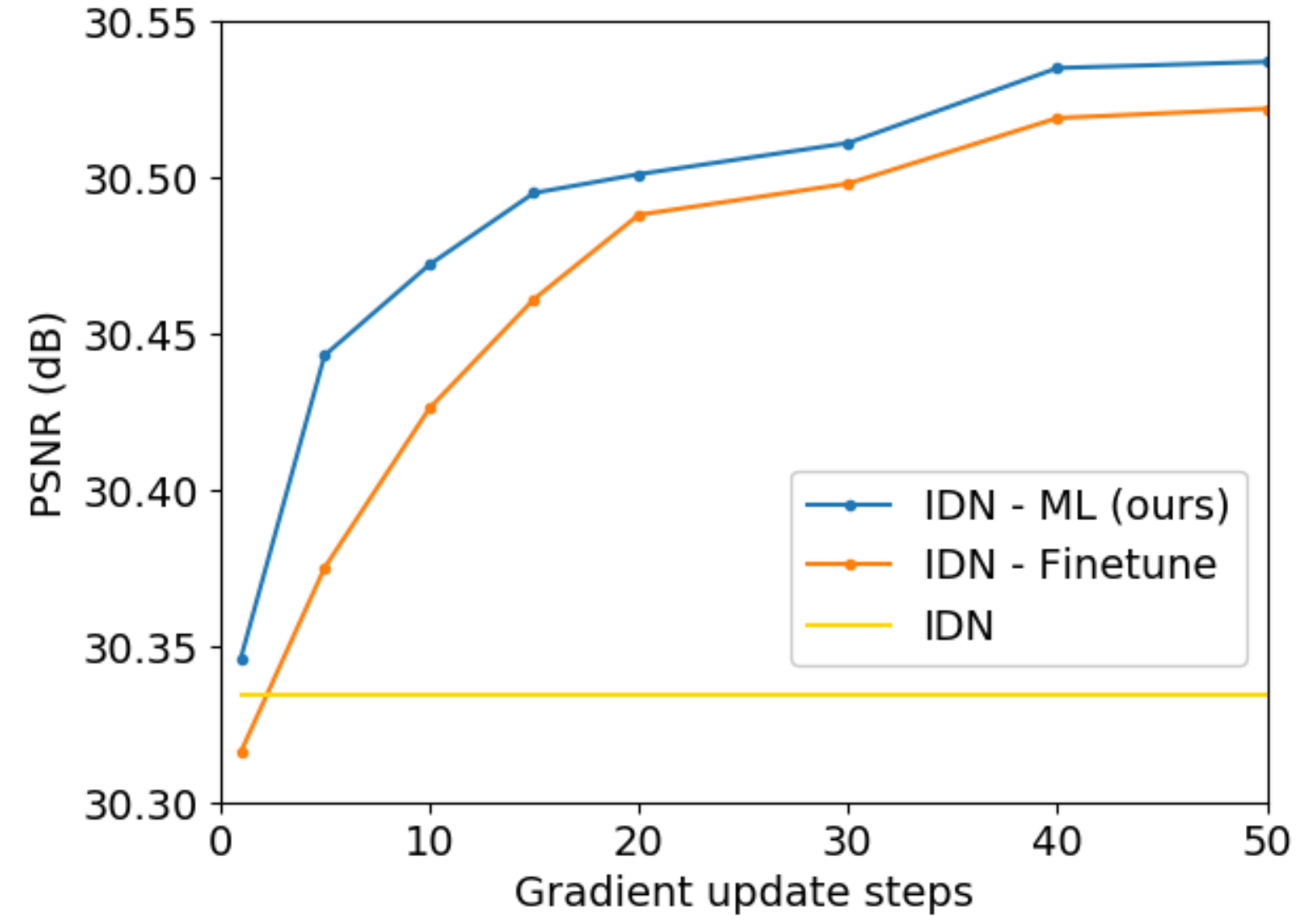}
\captionof{figure}{Performance changes of IDN-Finetune, and IDN-ML during adaptation on DIV2K.}
\label{fig_inner_update}
\end{minipage}%
\end{table}

First, we assume fixed bicubic SR kernel and compare PSNR and SSIM values of our SR networks on Urban100 and BSD100~\cite{martin2001database} datasets. For the comparison on DIV2K, test set of DIV2K is used since our networks are trained with DIV2K train set. 
Results are shown in~\tabref{table_mlsr_overall}, 
and we can see that PSNR and SSIM values of SR networks with meta-learning are higher than the original ones. Notably, the performance gaps on Urban100 are significantly larger than on other datasets, as urban scenes in the dataset mainly include structured scene with lots of patch-redundancy~\cite{selfex}. 

\begin{figure}[t]
    \centering
    \begin{minipage}[c]{.49\linewidth}
        \centering
        \captionsetup{font=footnotesize, justification=centering}
        \begin{subfigure}[c]{\linewidth}
            \includegraphics[width=\linewidth, height=.6\linewidth]{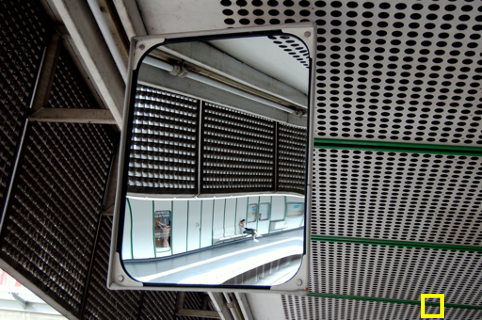}
        \end{subfigure}
        \newline
        \begin{subfigure}[c]{.185\linewidth}
            \includegraphics[width=\linewidth]{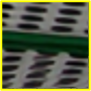}
            \caption*{IDN~\cite{idn}\\ \color{white} .}
        \end{subfigure}
        \begin{subfigure}[c]{.185\linewidth}
            \includegraphics[width=\linewidth]{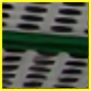}
            \caption*{IDN-ML\\(0)}
        \end{subfigure}
        \begin{subfigure}[c]{.185\linewidth}
            \includegraphics[width=\linewidth]{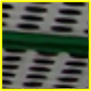}
            \caption*{IDN-ML\\(5)}
        \end{subfigure}
        \begin{subfigure}[c]{.185\linewidth}
            \includegraphics[width=\linewidth]{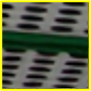}
            \caption*{IDN-ML\\(20)}
        \end{subfigure}
        \begin{subfigure}[c]{.185\linewidth}
            \includegraphics[width=\linewidth]{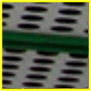}
            \caption*{GT\\\color{white} .}
        \end{subfigure}
    \end{minipage}
    \begin{minipage}[c]{.49\linewidth}
        \centering
        \captionsetup{font=footnotesize, justification=centering}
        \begin{subfigure}[c]{\linewidth}
            \includegraphics[width=\linewidth, height=.6\linewidth]{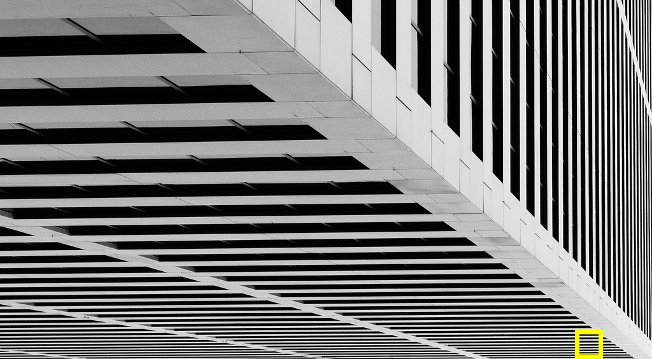}
        \end{subfigure}
        \newline
        \begin{subfigure}[c]{.185\linewidth}
            \includegraphics[width=\linewidth]{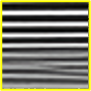}
            \caption*{IDN~\cite{idn}\\\color{white} .}
        \end{subfigure}
        \begin{subfigure}[c]{.185\linewidth}
            \includegraphics[width=\linewidth]{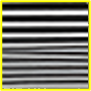}
            \caption*{IDN-ML\\(0)}
        \end{subfigure}
        \begin{subfigure}[c]{.185\linewidth}
            \includegraphics[width=\linewidth]{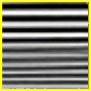}
            \caption*{IDN-ML\\(5)}
        \end{subfigure}
        \begin{subfigure}[c]{.185\linewidth}
            \includegraphics[width=\linewidth]{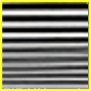}
            \caption*{IDN-ML\\(20)}
        \end{subfigure}
        \begin{subfigure}[c]{.185\linewidth}
            \includegraphics[width=\linewidth]{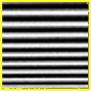}
            \caption*{GT\\\color{white} .}
        \end{subfigure}
    \end{minipage}
    \caption{Qualitative comparison with differently trained IDN~\cite{idn}. Number inside the bracket indicates the number of gradient update steps in run-time.}
    \label{figure_urban}
\end{figure}

To further explore the patch-recurrence of natural images, we train the networks with the meta-learning scheme in~\algoref{algorithm_mlsr} on the Urban100 dataset which includes a large number of similar patches. For meta-learning with Urban100, we use 50 images for training, 10 images for validation, and the remaining 40 images for test.
In~\tabref{table_urban}, we evaluate differently trained IDNs, and IDN trained on Urban100 with meta-learning algorithm outperforms other models.
Note that PSNR value of meta-learned IDN (IDN-ML) is relatively low before the adaptation, but improves dramatically with only 5 gradient updates (0.3dB gain). This proves that our MLSR method can learn better on images with rich patch-recurrence in urban scenes. More qualitative comparison results are shown in~\figref{figure_urban}, and the test images are particularly well restored with our network trained with meta-learning algorithm since specific patterns are repeated over the image itself. Moreover, we can see that the adapted parameters with more gradient update steps render visually much better results.

Moreover, in~\figref{fig_inner_update}, we show how PSNR value changes when the number of gradient steps in~\eqref{eq_update_inner} increases during meta-learning and test phases. As shown, our meta-learned model (IDN-ML) can quickly adapt SR parameters at test time, and achieves competitive results with only few gradient updates. Indeed, only 5 gradient updates can produce results which can be obtainable with $\sim$15 iterations of IDN with naive fine-tuning (IDN-Finetune).


\subsection{MLSR with unseen SR kernel}

In this section, we further conduct experiments to see the capability of the proposed MLSR algorithm in dealing with new and unseen SR kernel during the test phase. We carry out meta-learning in Algorithm~\ref{algorithm_mlsr} with randomly generated 5$\times$5 SR kernels on the DIV2K dataset and train for 30k iterations. Moreover, We generate 40k 5$\times$5 SR kernels as in \cite{chakrabarti2016neural}, and use 38k kernels for training, 1k for validation, and 1k for test.

\begingroup
\setlength{\tabcolsep}{4.5pt}
\renewcommand{\arraystretch}{1.1}
\begin{table}[t]
    \centering
    \begin{tabular}{lcccc}
    \hline
    Model & Set5          & Set14         & BSD100        & Urban100       \\ \hline
    ZSSR~\cite{zssr}                 & 29.68          & 27.76          & 27.53          & 25.02          \\
    IDN-ML / 0           & 28.10          & 26.22          & 26.19          & 23.48          \\
    IDN-ML / 5           & 29.17          & 27.08          & 26.82          & 24.36          \\
    IDN-ML / 20          & 29.86          & 27.67          & 27.32          & 24.96          \\
    IDN-ML / 100         & \textbf{30.41} & \textbf{28.12} & \textbf{27.75} & \textbf{25.42} \\ \hline
    \end{tabular}
    \caption{Comparing ZSSR~\cite{zssr} and meta-trained IDN on non-bicubic SR kernel. Right side of the slash indicates the number of gradient update steps in run-time.}
    \label{table_unknown_zssr}
\end{table}
\endgroup

\begin{figure}[t]
    \centering
    \begin{minipage}[c]{.49\linewidth}
        \centering
        \includegraphics[width=\linewidth]{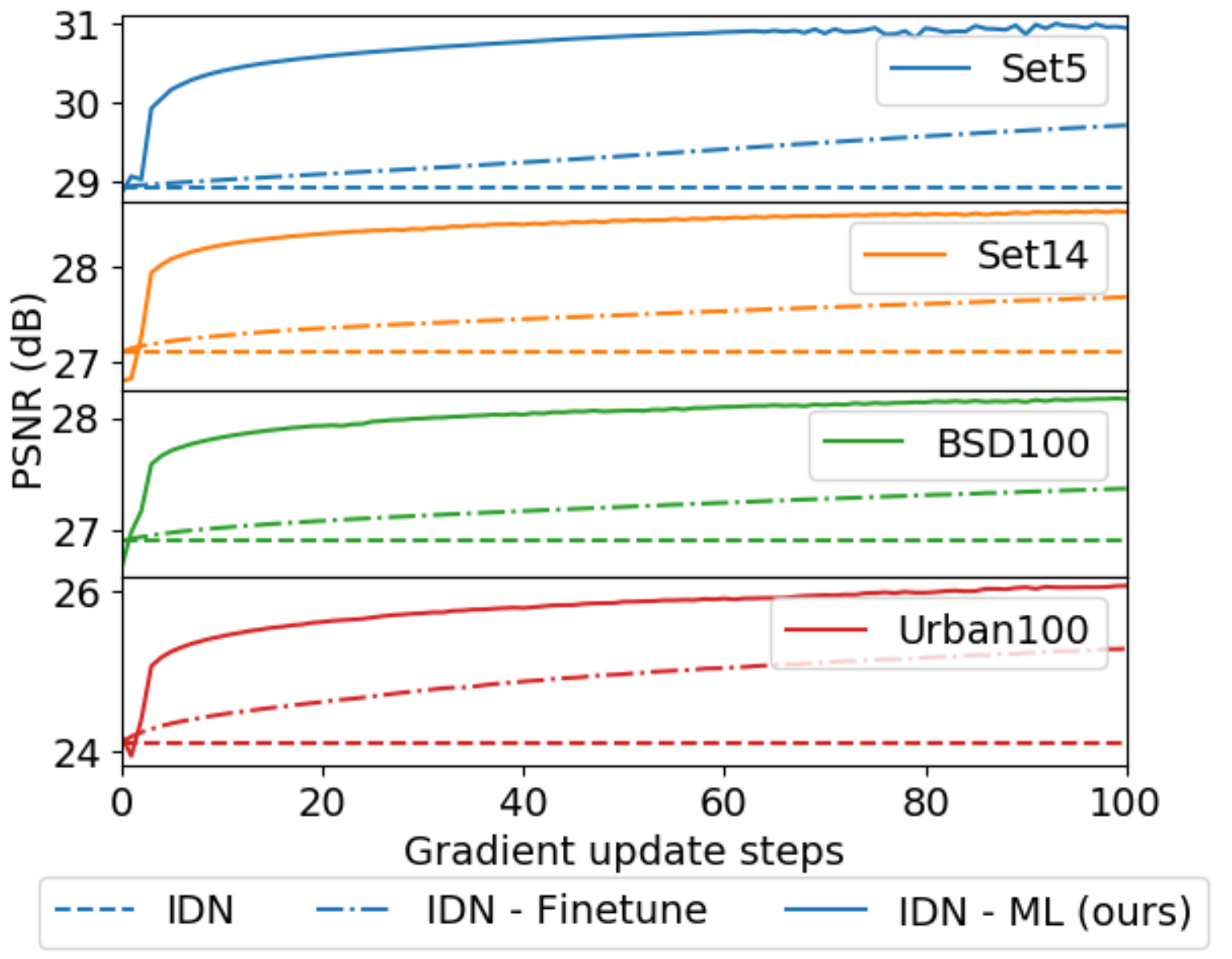}
        \caption{Performance curve of PSNR values on various test datasets. Random 5$\times$5 SR kernels are used.}
        \label{fig_unknown_plots}
    \end{minipage}
    \hfill
    \begin{minipage}[c]{.48\linewidth}
        \centering
        \includegraphics[width=\linewidth]{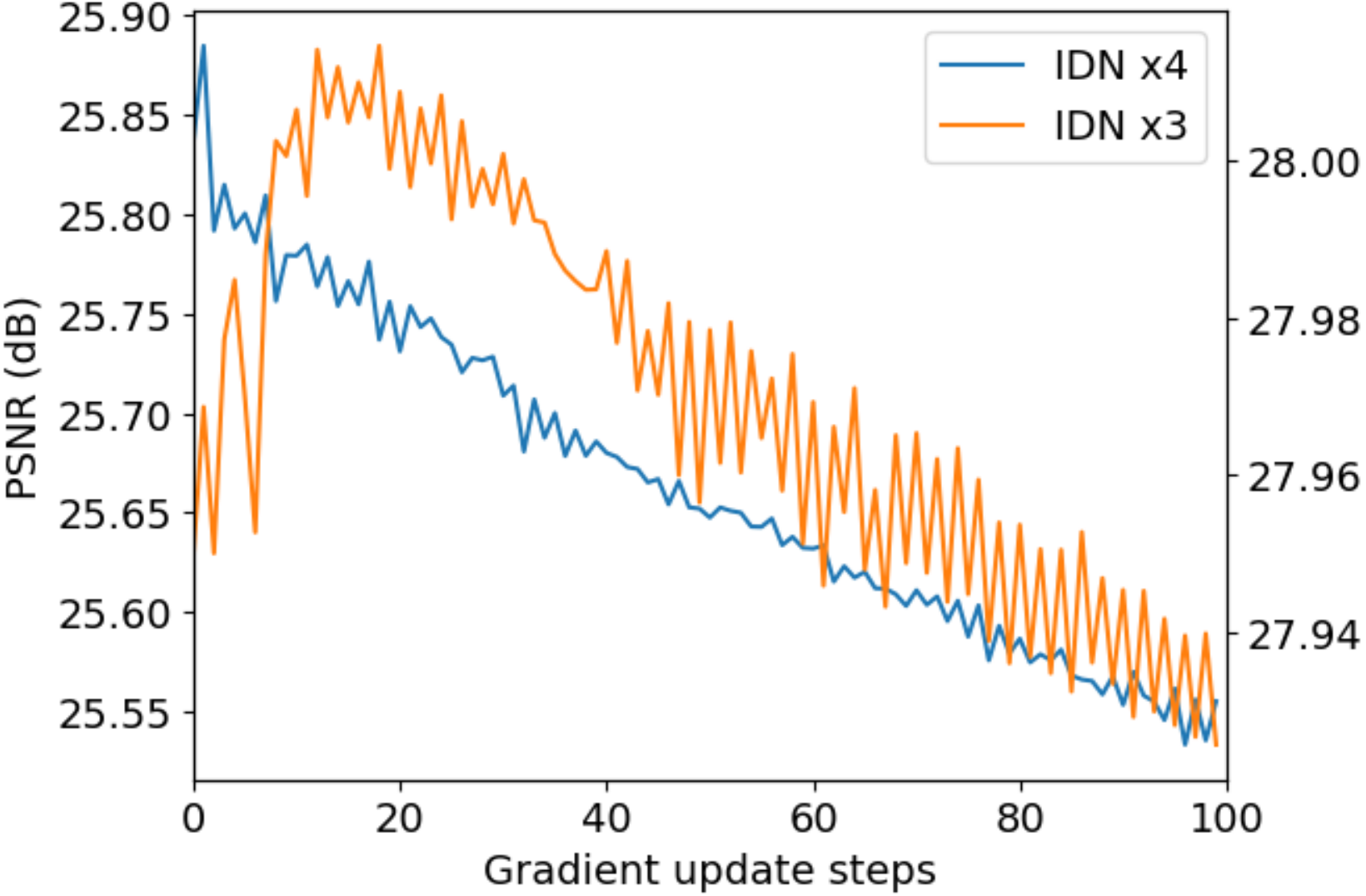}
        \caption{PSNR of IDN~\cite{idn} with scale $\times$3, $\times$4 on Urban100 dataset. Right and left sides of the y-axis indicate the PSNR values with respect to the upscaling factor $\times$3 and $\times$4, respectively.}
        \label{fig_sr_large_scale}
    \end{minipage}
\end{figure}

In~\tabref{table_finetune_unknown}, unlike fine-tuning with bicubic SR kernel, we need a large number of iterations ($\sim$1000) to achieve the highest PSNR value in dealing with non-bicubic SR kernel. However, our IDN-ML trained with many different SR kernels learnt the way to be quickly adapted to the new SR kernel given at test time, and it shows competitive results with only few gradient updates using the new kernel. Notably, we assume that the SR kernel is given or can be estimated with conventional methods as in~\cite{blindsr,zssr}. In \figref{fig_unknown_plots}, we can see that results with only 5 gradient updates (IDN-ML) are similar to the results from 350 iterations using the naive fine-tune without our meta-learning (IDN-Finetune).





After meta-training, we compare our model on Set5, Set14, BSD100 and Urban100 datasets. In the inference stage, an SR kernel that has not been shown during training stage, and an LR image degraded with that SR kernel are provided. The results in \figref{fig_unknown_plots} show consistent improvements for various datasets as the number of gradient update steps increases. 
Specifically, the performances raise strikingly ($\sim$1dB) at around 5 iterations, and it verifies that the network can quickly adapt to the given input image and SR kernel at test time with the small number of updates.
Notably, the result on Urban100 is slightly different from others. Performance of the adapted network on Urban100 improves more rapidly than adapted networks on other datasets as rich patch-recurrence with urban scenes helps to handle newly seen SR kernels at test time.

\begin{figure}[h]
    \centering
    \includegraphics[width=\linewidth]{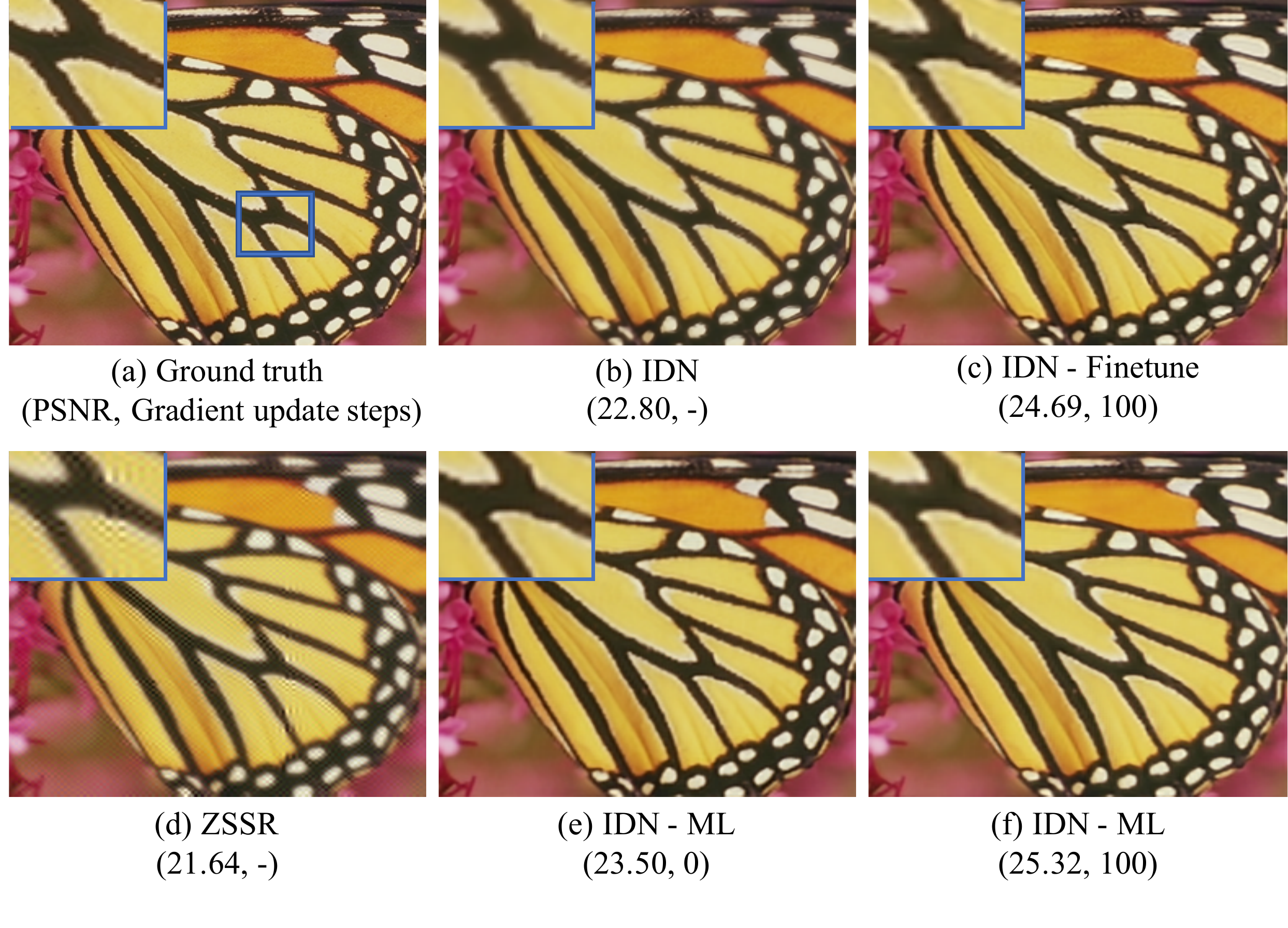}
    \caption{The ``butterfly" image from Set5 dataset with upscaling factor $\times$2. Input LR image is downscaled with a non-bicubic 5$\times$5 SR kernel. }
    \label{fig_updatesnknown_butterfly}
\end{figure} 
\begin{figure}[h]
    \centering
    \includegraphics[width=\linewidth]{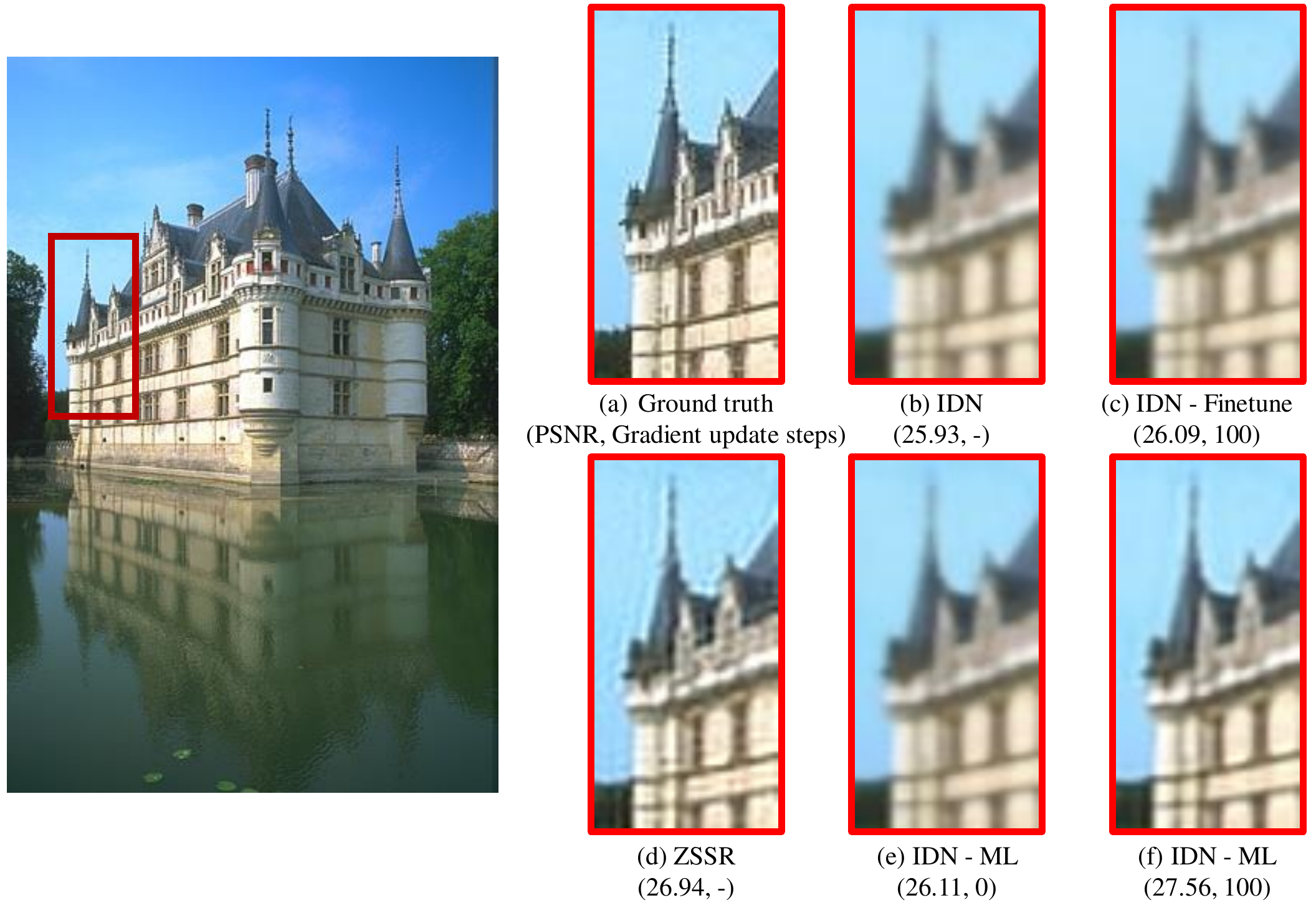}
    \caption{The ``102061" image from BSD100 dataset with upscaling factor $\times$2. LR image is generated using a non-bicubic kernel. Our method achieves better performance than naive fine-tuning with the same number of inner updates at run-time. }
    \label{fig_updatesnknown_bsd}
\end{figure} 

In~\tabref{table_unknown_zssr}, we also compare ours with ZSSR~\cite{zssr} on numerous dataset with SR kernels used in ZSSR, and our proposed method with 20 gradient updates shows competitive results compared to ZSSR, and significantly outperforms ZSSR when adapted for 100 iterations. Notably, ours with 100 iterations takes only 6 minutes to restore 100 urban images, but ZSSR requires more than 3 hours with GeForce RTX 2080Ti.

In \figref{fig_updatesnknown_butterfly} and \figref{fig_updatesnknown_bsd}, we compare visual results by naive fine-tuning (IDN$-$Finetune), meta-learning (IDN$-$ML) and ZSSR.
We see that the quality improves significantly within few iterations with our MLSR algorithm, and the boundaries are restored gradually as iteration goes. Moreover, artifacts near boundaries caused by ZSSR are not produced by our proposed method.

\subsection{SR with large scaling factor}

Finally, we study the validity of the patch-recurrence property with large SR scaling factor. Unfortunately, as shown in \figref{fig_sr_large_scale}, exploiting patch-recurrence on big scale factor is hard
(\ie $\times$3 or $\times$4). Maximal performance gained by fine-tuning with large SR factors are around 0.04dB which are negligible. Therefore, to produce large images with large scaling factors, we can employ multi-scale (coarse-to-fine) approaches embedded into the conventional SR methods with small scale factor (\eg $\times$1.25) which also exploit self-similarity nature of the given test images~\cite{glasner,selfex,zssr}.


\section{Conclusion}
In this work, we introduced a new SR method which utilizes both the power of deep learning with large external dataset and additional information available from the input image at test time. To this end, we proposed a novel Meta-Learning for SR (MLSR) algorithm which enables quick adaptation of SR parameters using only input LR image during the test phase. MLSR can be combined with conventional SR networks without any architecture changes, and can utilize the patch-recurrence property of the natural image, which can further boost PSNR performance of various deep learning-based methods. In addition, MLSR can handle non-bicubic SR kernel that exists in real world because meta-learned networks can be adapted to the specific input image. In experiments, we show that our MLSR can greatly boost up the performance of existing SR networks, with a few gradient update steps. Moreover, we experimentally demonstrated that MLSR takes advantage of the patch-recurrence well, by showing the performance improvements on the Urban100 dataset, where patch-recurrence occurs frequently. Finally, the proposed MLSR algorithm was also validated with the unseen non-bicubic SR kernel and showed that MLSR required less gradient updates than naive fine-tuning. We believe that the proposed method can be applied not only to SR but also to various types of reconstruction and low-level vision tasks.


%
%
\bibliographystyle{splncs04}
\bibliography{egbib}
\end{document}